\pdfoutput=1
\documentclass[preprint,10pt,authoryear]{elsarticle}

\usepackage{mathptmx}                       % Times text and math (newtxtext not installed)
\usepackage[a4paper,top=4.3cm,right=4.8cm,bottom=4.3cm,left=4.8cm]{geometry}
\usepackage{setspace}
\usepackage{amsmath,amssymb}
\usepackage{booktabs,array,multirow,longtable}
\usepackage{rotating}                       % sidewaystable for tables wider than the 11.4 cm text block
\usepackage[font={footnotesize,stretch=1},labelsep=period]{caption}   % captions 8pt, single spaced
\usepackage{etoolbox}
\AtBeginEnvironment{table}{\setstretch{1}\normalsize}            % table bodies 10pt, single spaced
\AtBeginEnvironment{sidewaystable}{\setstretch{1}\normalsize}
\usepackage[section]{placeins}
\usepackage[hidelinks]{hyperref}

\newif\ifrefsdoublespaced \refsdoublespacedtrue
\AtBeginEnvironment{thebibliography}{\ifrefsdoublespaced\else\setstretch{1}\normalsize\fi}
\makeatletter
\def\ps@pprintTitle{\let\@oddhead\@empty\let\@evenhead\@empty\def\@oddfoot{\hfil\thepage\hfil}\let\@evenfoot\@oddfoot}
\makeatother


\journal{Pattern Recognition}

\newcommand{\tless}{T-LESS}
\newcommand{\dgs}{3DGS}
\newcommand{\fgip}{FGIP}

\begin{document}

\begin{frontmatter}

\title{What Do Scan-Derived Class Prototypes Add? Disentangling Supervision, Prototype Content and Query Protocol in Recognition over Frozen Foundation Features}

\author[gt]{Chenxi Tao}
\ead{ctao40@gatech.edu}

\author[kitech]{Hong-In Won}

\author[gt]{Seung-Kyum Choi\corref{cor1}}
\ead{schoi@me.gatech.edu}
\cortext[cor1]{Corresponding author}

\affiliation[gt]{organization={George W. Woodruff School of Mechanical Engineering},
            addressline={Georgia Institute of Technology},
            city={Atlanta},
            state={GA},
            postcode={30332},
            country={USA}}

\affiliation[kitech]{organization={Korea Institute of Industrial Technology},
            city={Cheonan-si},
            state={Chungcheongnam-do},
            country={Korea}}

\begin{abstract}
A scan supplies labeled images and a geometric reference. We separate their contributions in a
recognizer whose scan-derived prototype matrix acts as a supervised head's fixed output layer.
On T-LESS, HOPE and 18 self-collected industrial parts, we test real, random and exactly permuted
prototypes, matched geometry-free classifiers, stronger appearance rules and paired background
protocols. Across DINOv2-giant and MetaCLIP-H with real-background queries, the largest
fused-accuracy advantage of the real prototypes over either control is one percentage point; larger
differences favor controls, by up to 2.8 points in arm means. On HOPE with DINOv2-giant the head
alone is 2.8 points above exact permutations (95\% interval: 0.8--4.7); this advantage does not
reach fusion and is not observed on MetaCLIP-H. On DINOv2-giant, matched logistic regression comes
within 0.5 points of fusion on T-LESS and exceeds it on HOPE and the self-collected parts.
Against white cutouts, real HOPE query backgrounds lower image-prototype accuracy by 43 points on
DINOv2-giant and 13 on MetaCLIP-H. The audit separates prototype content, label supervision and
query protocol.
\end{abstract}
\end{frontmatter}

\section{Introduction}\label{sec:intro}

A short object scan supplies both images labeled by object identity and a three-dimensional
reference. These two products offer different explanations for improved recognition: the images
can train a classifier, while the reference can encode the object's shape. Separating their
contributions matters when CAD models are unavailable and a reconstruction serves as the
reference~\citep{tao2026prior,sun2022onepose,wen2024foundationpose,gfreedet2024}. Frozen foundation features~\citep{oquab2024dinov2} and fast
reconstruction with 3D Gaussian Splatting (\dgs)~\citep{kerbl2023gaussian} make such systems
practical, but an improvement over image-prototype matching alone does not identify which product
of the scan supplied it.

We examine this distinction in a recognizer that maps frozen image features to scan-derived class
prototypes. A trainable head predicts a vector whose similarities to the prototypes give the class
scores. The prototype matrix therefore acts as the head's fixed output layer; cross-entropy trains
the preceding mapping using the onboarding labels. The query is an RGB crop, and geometry enters
through the prototype arrangement. Existing results on fixed output layers and predefined
prototypes motivate testing both arbitrary prototype matrices and meaningful class
assignments~\citep{hoffer2018fix,mettes2019hpn}.

We ask whether the scan-derived matrix contributes beyond matched label supervision,
whether any contribution survives a stronger appearance representation, and how the query
background affects those conclusions. We test real, random and exactly class-permuted prototype matrices;
geometry-free classifiers on the same optimization subsets; global and patch-level appearance
matching; and paired white-cutout and real-background queries. The evaluation covers \tless, HOPE and
self-collected industrial parts (\fgip); the prototype audit runs on DINOv2-giant and is repeated on
MetaCLIP-H~\citep{xu2024metaclip}.

We provide a reproducible audit for recognizers whose fixed class-prototype matrices are read by a
class-supervised head. It tests whether the original shape-to-class assignment adds predictive
value beyond matched supervision and arbitrary prototypes. It then checks the effect of stronger
appearance matching and paired query protocols. In the tested system, most fused gains persist
without the original assignment, while a conditional head-level advantage remains on HOPE with
DINOv2-giant. Separating head and fused outcomes lets readers assess whether the contribution of a
prototype matrix improves the final recognizer under matched conditions. Section~\ref{sec:discuss}
assembles the interventions and reporting requirements into a procedure others can reproduce.
An external experiment applies the assignment intervention to a published semantic-prototype
recognizer with full-network training (Section~\ref{sec:hpn}).

\section{Related Work}\label{sec:related}

\paragraph{How reference geometry enters recognition}
The BOP benchmarks distinguish seen-object, unseen-object and model-free
settings~\citep{hodan2020bop,hodan2024bop}. CNOS, SAM-6D and FoundPose use rendered references with
frozen descriptors~\citep{nguyen2023cnos,lin2024sam6d,ornek2024foundpose}; MegaPose and GigaPose use
learned render-and-compare or template matching~\citep{labbe2022megapose,nguyen2024gigapose}.
FoundationPose, OnePose and GFreeDet obtain references from images, video or reconstructed
twins~\citep{wen2024foundationpose,sun2022onepose,gfreedet2024}. FreeZe registers query and model
point clouds using frozen geometric and visual features~\citep{caraffa2024freeze}, while ULIP and
OpenShape learn joint image--text--point-cloud embeddings for cross-modal
retrieval~\citep{xue2023ulip,liu2023openshape}. Table~\ref{tab:scope} separates these reference uses
from a per-class matrix fitted through a supervised head. This distinction determines which
intervention can test the role of geometry.

\paragraph{Scans, descriptors and frozen features}
Neural radiance fields and \dgs{} reconstruct objects from posed
images~\citep{mildenhall2020nerf,kerbl2023gaussian}. GFreeDet, a BOP challenge report, uses rendered
Gaussian-Splatting templates for matching~\citep{gfreedet2024}; our system uses the reconstruction
to obtain point-cloud descriptors. Global shape signatures and learned point encoders provide
alternative summaries~\citep{osada2002shape,qi2017pointnet,qi2017pointnetpp,yu2022pointbert}.
On the image side, self-supervised transformers~\citep{caron2021dino,oquab2024dinov2,simeoni2025dinov3}
and image--text encoders~\citep{radford2021clip,zhai2023siglip,tschannen2025siglip2,xu2024metaclip}
supply frozen representations. Image compositing~\citep{dwibedi2017cutpaste} and texture
randomization~\citep{honig2024shapebiased} instead alter the training images. We compare mean image
prototypes~\citep{snell2017prototypical} with stronger appearance rules while keeping the feature
extractor frozen.

\paragraph{Fixed outputs and attribution controls}
Fixing a classifier's output layer to random orthogonal or Hadamard weights can retain accuracy
when the preceding network is trained~\citep{hoffer2018fix}. Hyperspherical Prototype Networks (HPN) show
that predefined prototypes can also benefit from class-semantic structure~\citep{mettes2019hpn}.
Random class-to-codeword assignments were studied in error-correcting output
codes~\citep{dietterich1995ecoc} and in hyperspherical prototypical
learning~\citep{lindstrom2024coding}. These precedents motivate our controls. We apply them under
frozen features and an onboarding-to-query domain shift to test whether a scan-derived target
supplies shape evidence or serves as a class code.

\paragraph{Representation and protocol}
Strong embeddings with simple classifiers challenge apparent gains in few-shot
learning~\citep{chen2019closer,tian2020rethinking}; fair evaluation likewise changes metric-learning
comparisons~\citep{musgrave2020reality}. Background dependence and environmental shifts affect
recognition~\citep{xiao2021backgrounds,beery2018terra,barbu2019objectnet}. Our paired query analysis
isolates a specific protocol change: replacing all pixels outside the visible-object mask,
including neighbors and occluders, while retaining the crop and target identity.

\begin{table}[htbp]
\centering
\caption{Reference geometry and object-specific supervision. A class permutation directly tests prototype assignment in the final row.}
\label{tab:scope}
\setlength{\tabcolsep}{3pt}
\begin{tabular}{>{\raggedright\arraybackslash}p{66pt}>{\raggedright\arraybackslash}p{101pt}>{\raggedright\arraybackslash}p{133pt}}
\toprule
Method & Query and reference & Decision path / object-specific training \\
\midrule
CNOS, SAM-6D & RGB; CAD-rendered templates. SAM-6D also uses depth for pose. & Frozen template matching; none. \\
FoundPose & RGB; CAD RGB-D templates. & Template retrieval, 2D--3D correspondences and PnP; none. \\
GFreeDet & RGB; Gaussian-Splatting-rendered templates. & Frozen template matching; none. \\
FreeZe & RGB-D; model point cloud with frozen geometric and visual features. & Registration of query and model clouds; none. \\
ULIP, OpenShape & Point cloud or image; joint image--text--point-cloud embedding. & Retrieval in the joint embedding; none per object. \\
Audited system & RGB; per-class $45$-D scan descriptors ($G$). & $G$ is a fixed classifier output layer; head $h$ learns from labeled onboarding frames. \\
\bottomrule
\end{tabular}
\end{table}

\section{System and audit design}\label{sec:method}

\begin{figure}[htbp]
\centering
\includegraphics[width=\linewidth]{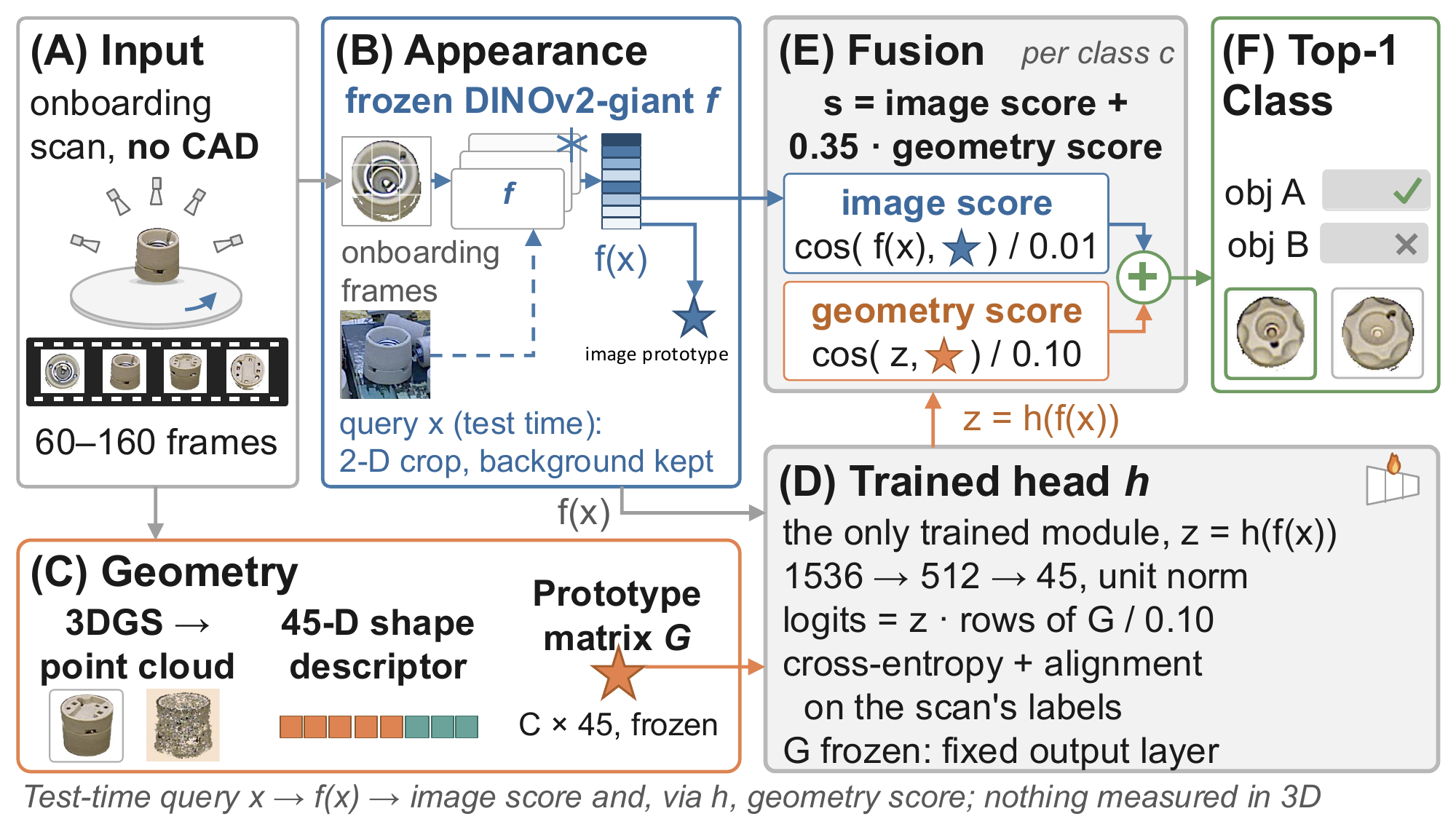}
\caption{The audited computation. (A) Labeled onboarding scan of $60$--$160$ frames per object, without CAD. (B) Frozen image encoder $f$ and one mean image prototype per class (blue star); the query $x$ is an RGB crop with its background kept. (C) \dgs{} reconstruction, point cloud and $45$-D descriptor per class, stacked into the fixed prototype matrix $G$ (orange star). (D) The head $h$, the only trained module, fitted on the scan labels with logits $z^\top g_c/0.10$. (E) Fusion of the image score with $0.35$ times the geometry score. (F) Top-1 class. The controls act at $G$ (Sections~\ref{sec:codebook} and~\ref{sec:backbone2}), on lanes C--E (Section~\ref{sec:supervision}), on the image prototype (Section~\ref{sec:appearance}), on the query crop (Section~\ref{sec:protocol}) and on $f$ (Section~\ref{sec:backbone2}).}
\label{fig:pipeline}
\end{figure}

The audited system is a closed-set recognizer with two branches over one frozen image encoder.
One branch is a mean image prototype per class; the other is a trained head whose output is scored
against a fixed scan-derived prototype matrix $G$. Figure~\ref{fig:pipeline} shows both branches
and the points at which the controls act.

\subsection{Data and acquisition}\label{sec:datasets}
Table~\ref{tab:datasets} summarizes the acquisition. \tless~\citep{hodan2017tless} contains textureless industrial parts and uses
\texttt{train\_primesense} onboarding frames and \texttt{test\_primesense} queries;
objects $19$ and $20$ were excluded when reconstruction failed, before recognition evaluation.
HOPE~\citep{tyree2022hope} contains textured household objects and uses \texttt{onboarding\_static} and validation queries.
\fgip{} contains two gears, eight nuts and eight screws: onboarding comes from video A and
queries from an independently captured video B.
All onboarding crops are segmented foregrounds.

\paragraph{Query protocols}
Queries are scored under two protocols. The real-background protocol takes the raw crop of the
ground-truth box and keeps its scene content; it is the main protocol and applies wherever no
protocol is named. The white-cutout protocol pastes the foreground selected by the ground-truth
visibility mask onto a white canvas, which removes the background, the neighboring objects and the
occluding objects inside the box. On \fgip{} the mask and the box come from SAM2
segmentation of the test video~\citep{ravi2025sam2}. On HOPE both protocols use the same $915$
boxes, so the two query sets are paired one to one. On \fgip{} both use the same $1{,}927$ frames,
and the white cutout removes only the turntable surface. \tless{} queries are scored with real
backgrounds only.

\begin{table}[htbp]
\centering
\caption{Datasets. Onboarding uses foreground crops; queries retain the scene within the object box. FGIP queries come from an independent video.}
\label{tab:datasets}
\setlength{\tabcolsep}{3pt}
\begin{tabular}{lrrr}
\toprule
Dataset (classes) & Onboarding & Per class & Queries \\
\midrule
\tless{} ($28$) & $4{,}480$ & $160$ & $6{,}395$ \\
HOPE ($28$) & $1{,}680$ & $60$ & $915$ \\
\fgip{} ($18$) & $2{,}839$ & $\le160$ & $1{,}927$ \\
\bottomrule
\end{tabular}
\end{table}

\begin{figure}[htbp]
\centering
\includegraphics[width=\linewidth]{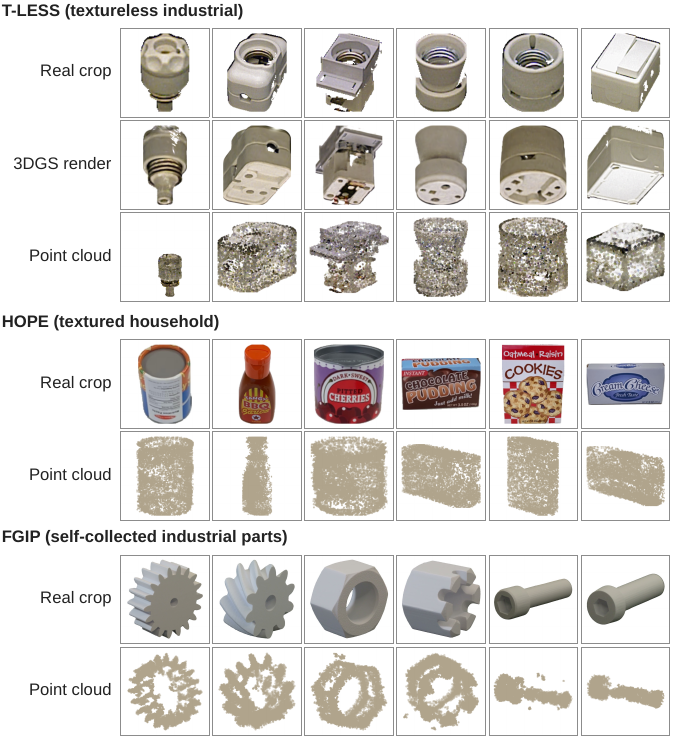}
\caption{Onboarding references for six objects each of \tless{} (top), HOPE (middle) and \fgip{} (bottom). \tless{} rows: a segmented onboarding crop with the foreground on white (labeled ``real crop''), a novel view rendered from the \dgs{} reconstruction, and the extracted point cloud. HOPE rows: a segmented onboarding crop and the extracted point cloud. \fgip{} rows: a segmented frame of the onboarding video on white and the extracted point cloud. The \fgip{} block shows two gears, two nuts and two screws of the $18$ parts: gears A and B, nuts A and B, screws A and B. \tless{} clouds carry the Gaussian colors; the other clouds are drawn in one color. All images are onboarding data. The point cloud is the only reconstruction product that enters the class descriptors (Section~\ref{sec:branch}).}
\label{fig:recon}
\end{figure}

\paragraph{Reconstruction}
Nerfstudio \texttt{splatfacto-big} trains each object on a white background for $20{,}000$ iterations
on \tless{} and $15{,}000$ on HOPE, with point-cloud initialization on both. BOP \texttt{scene\_gt} rotations and translations are converted
to camera-to-world poses; \fgip{} instead uses COLMAP poses. \tless{} initialization uses onboarding
depth. On \tless{} and HOPE, exported Gaussian centers are cropped to $0.15$\,m from the origin and filtered with
Open3D statistical outlier removal ($20$ neighbors, standard-deviation ratio $2.0$).
\fgip{} replaces this metric crop with opacity $\ge0.01$, median centering, a radius of $1.5$
times the $90$th-percentile radius, and three outlier-removal passes. Settings are fixed within each dataset.
Figure~\ref{fig:recon} shows the resulting references for the three datasets, including residual floaters
associated with sparse-view reconstruction~\citep{xiong2023sparsegs}.

Table~\ref{tab:cost} reports measured reconstruction cost. Training takes $134$--$156$\,s per
object; preparation and export add a few seconds. The attribution controls of
Section~\ref{sec:setup} take about $10$ GPU minutes per run including feature extraction: $135$
DINOv2 and $45$ MetaCLIP runs total about $30$ GPU hours. Logistic probes take seconds on a CPU.

\begin{table}[htbp]
\centering
\caption{T-LESS reconstruction cost per object ($28$ objects, \texttt{splatfacto-big}, $20$k iterations, one NVIDIA RTX 5000 Ada). Each fixed-elevation ring contains $32$ azimuth views.}
\label{tab:cost}
\setlength{\tabcolsep}{3pt}
\begin{tabular}{lccc}
\toprule
Coverage & Views & Minutes & GB \\
\midrule
One equatorial ring & $32$ & $2.41$ & $0.14$ \\
Three rings & $96$ & $2.66$ & $0.15$ \\
Five rings (main) & $160$ & $2.83$ & $0.17$ \\
\bottomrule
\end{tabular}
\end{table}

\subsection{Class descriptors and prediction}\label{sec:branch}
Each cloud is subsampled to at most $12{,}000$ points, mean-centered and scaled by its maximum
radius. Its $45$-dimensional descriptor contains three box edge-length ratios, three covariance
eigenvalue ratios, three box statistics (product, mean and standard deviation of extents), and
three $12$-bin histograms: radial distance in the first two principal coordinates, height in
the third normalized to its span, and pairwise distance. The last uses all $2{,}080$ pairs among $65$ seeded
sampled points ($\lfloor\sqrt{4096}\rfloor+1$). Standardizing each dimension across classes and
L2-normalizing each row gives the prototype matrix $G\in\mathbb{R}^{C\times45}$, whose row $g_c$ is the
geometry prototype of class $c$.

Centering and scaling normalize translation and scale. Eigenvalue ratios and pairwise distances
are rotation invariant. The radial histogram is invariant to eigenvector signs and to ambiguity
when the first two eigenvalues coincide; a sign flip of the third eigenvector reverses the height
histogram. When the second and third eigenvalues coincide, both histograms can change. The box is
axis-aligned in the input frame, so its six terms depend on pose: the implemented descriptor is
not exactly SE(3)-invariant. Sampling also changes $G$: mean row-wise cosines between seed-specific
real matrices range from $0.75$--$0.84$ on \tless, $0.85$--$0.92$ on HOPE and $0.95$--$0.97$ on
\fgip.

Frozen DINOv2-giant supplies L2-normalized $1536$-D CLS features $f(x)$. The image prototype
$p_c$ is the normalized mean of class $c$'s onboarding features. A head
$h:1536\!\rightarrow\!512\!\rightarrow\!45$ uses ReLU, dropout $0.1$ and L2-normalized output
$z(x)$. For class $c$,
\begin{align}
\ell_c(x)&=z(x)^\top g_c/0.10,\label{eq:logits}\\
L(x,y)&=\operatorname{CE}(\ell(x),y)+0.25\bigl[1-z(x)^\top g_y\bigr],\label{eq:loss}\\
s_c(x)&=f(x)^\top p_c/0.01+\lambda\ell_c(x).\label{eq:fusion}
\end{align}
Only $h$ is trained; the image encoder, $p_c$ and $G$ remain fixed (Fig.~\ref{fig:pipeline}).
AdamW uses learning rate $10^{-3}$, weight decay $10^{-4}$, batch size $64$ and $80$ epochs.
For each seed, $20$ onboarding frames per class are held out and select the best-accuracy epoch.
Seeds control initialization, shuffling, the split and point sampling. The audited fusion uses
$\lambda=0.35$; image-only uses $\lambda=0$, while head-only uses $\ell$. The fusion weight and the
patch-matcher settings of Section~\ref{sec:setup} were fixed before the controls were run.

Equation~\ref{eq:logits} makes $G$ a fixed output layer trained through class labels. Prototype
angles can affect margins, gradients and generalization. An orthogonal coordinate change applied
jointly to $z$ and $G$ preserves the logits and alignment loss, whereas arbitrary replacement of
$G$ need not.

\subsection{Interventions and statistical units}\label{sec:setup}
The real arm uses five seed-specific matrices $G_s$ and head seeds $s=0,\ldots,4$.
The random arm generates five independent prototype matrices (generator seeds $1$--$5$) from randomly
rotated ellipsoidal shells with Gaussian blobs, using identical descriptor preprocessing and
three head seeds per matrix. The two $28$-class datasets share these random matrices because
generation depends on seed and class index.
The permutation arm uses five fixed-point-free row permutations per real matrix and retrains on
the matched seed and split: $G'_{sk}=P_kG_s$. All $75$ saved matrices are bitwise exact row
reorderings. Recomputing descriptors after reassignment would resample the clouds; matrices built
that way have mean row cosines of $0.78$--$0.97$ to $P_kG_0$, so the permutations reorder the
saved rows.

Random replacement tests whether object-derived rows are necessary but also changes conditioning.
Permutation isolates class assignment while preserving $G^\top G$, singular values and the
multiset of pairwise row cosines, since
$G'G'^\top=P_kGG^\top P_k^\top$.
As a positive control, permuting directly matched image prototypes reduces accuracy from
$0.560$, $0.399$, $0.803$ to $0.013$, $0.021$, $0.010$ on \tless, HOPE and \fgip{}, respectively
(chance: $0.036$, $0.036$, $0.056$). Thus the intervention disrupts a reference whose class
assignment is directly read at inference.

For real versus permuted, define accuracy $a(\cdot)$ and the paired difference
\begin{equation}
\Delta_s=a(G_s)-\frac15\sum_{k=1}^{5}a(P_kG_s).
\end{equation}
We summarize this over five seeds with a $95\%$
$t$-interval. We also inspect differences over permutation means and the $25$ paired cells;
disagreement between the two intervals calls for the more conservative conclusion.
For random controls the unit is one random prototype matrix, with its accuracy averaged over three
seeds. The $95\%$ $t$-interval of the real-minus-random difference uses five matrices, four degrees of freedom, and a standard error combining between-matrix
variance with seed variance of the real mean. Run-level Welch statistics are descriptive because
shared matrices induce dependence. We run no equivalence test. A difference whose interval includes
zero is reported as not detected with these runs. Fused-level differences between the real prototype matrix
and its controls are reported with their intervals; statements of the largest advantage refer to
observed arm means.

\paragraph{Geometry-free references}
Logistic regression uses fixed $C=1$ on the same frozen features and labels. The softmax head
uses $1536\!\rightarrow\!512\!\rightarrow\!C$ with the same optimization schedule as $h$.
Matched references use each seed's optimization subset and the same held-out checkpoint rule for
the softmax head. Full-budget references use all onboarding frames and the softmax head's final
epoch. Logistic regression has no checkpoint selection and is deterministic for a fixed subset;
softmax results use five seeds.

\paragraph{Appearance references}
Five global rules use mean prototypes, the best template, the mean of the best five,
eight $k$-means sub-prototypes, or temperature-weighted soft voting. Patch bag-of-words (BoW)
uses cosine similarity between TF--IDF histograms, modeled on FoundPose's retrieval stage~\citep{ornek2024foundpose}, without
correspondence estimation or PnP. Its $2{,}048$-word vocabulary is fitted for $25$ $k$-means
iterations (seed $0$), using at most $40$ onboarding frames per class and $200{,}000$ tokens.
We test final-layer and hidden-state-$30$ features (of $40$), with best-template or best-five
matching over all onboarding frames. A patch with mean intensity above $0.93$ is discarded;
queries retaining fewer than four patches instead keep all $256$. No semantic mask is applied by
the matcher; $68\%$ of real-background HOPE queries retain all patches.
The rules require more storage and query computation than a mean prototype. Each rule calibrates
its fusion weight on held-out onboarding frames over $[0,2]$.

OpenAI Codex (September 2026) and Anthropic Claude (September 2026) assisted the implementation of
the experiment and analysis scripts; every reported number is recomputed from the stored run outputs.

\section{Attribution results}\label{sec:exp}

\subsection{Prototype content and class assignment}\label{sec:codebook}
\begin{figure}[htbp]
\centering
\includegraphics[width=\linewidth]{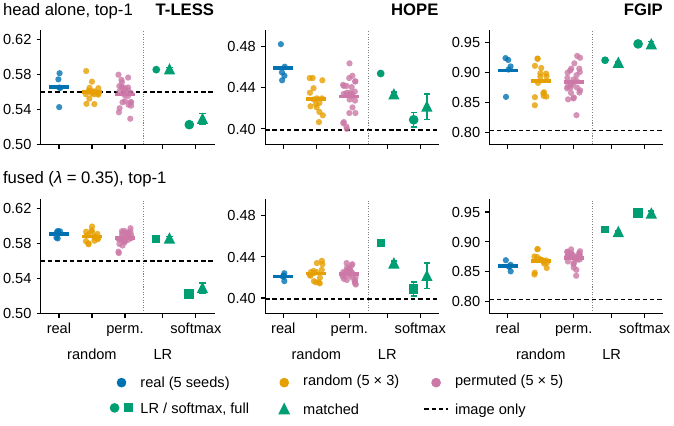}
\caption{Prototype controls on DINOv2-giant with real-background queries, for \tless{}, HOPE and \fgip{} (columns). Top row: top-1 accuracy of the head alone; bottom row: fused accuracy at $\lambda=0.35$. Each point is one training run and each bar the mean of its arm: real prototypes (blue, $5$ seeds), random prototypes (orange, $5$ matrices $\times$ $3$ seeds) and exact row permutations $P_kG_s$ (purple, $5$ permutations $\times$ $5$ seeds). Green markers right of the dotted line are geometry-free references on the same features and labels, repeated in both rows: logistic regression (LR) and softmax head trained on all onboarding frames (circle, square) and on the matched subset (triangles; mean and SD over $5$ seeds). The dashed line is image-only accuracy.}
\label{fig:codebook}
\end{figure}

The real head and fusion improve upon mean image prototypes (Table~\ref{tab:codebook};
Fig.~\ref{fig:codebook}). Tables~\ref{tab:codebook} and~\ref{tab:codebookdiff} separate arm
accuracies from their differences. Under fixed fusion, the random and permuted prototype matrices
retain the fused improvement to within $0.5$ points, which is the largest advantage of the real
matrix on DINOv2-giant (\tless{}, permuted). \fgip{} instead favors permutation by $1.3$ points, with $21$
of $25$ paired cells negative.

\begin{table}[htbp]
\centering
\caption{DINOv2-giant: audited system and prototype controls on real-background queries (top-1, mean$\pm$SD). Real: $5$ head seeds; random: $5$ prototype matrices $\times$ $3$ head seeds; permuted: $5$ exact row permutations $\times$ $5$ head seeds. Fusion uses $\lambda=0.35$; best\_val is held-out onboarding accuracy.}
\label{tab:codebook}
\setlength{\tabcolsep}{2pt}
\begin{tabular}{llccc}
\toprule
Dataset & Output & Real & Random & Permuted \\
\midrule
\tless{} & Image-only & \multicolumn{3}{c}{$0.560$} \\
 & head & $0.566{\pm}0.015$ & $0.560{\pm}0.009$ & $0.557{\pm}0.013$ \\
 & fused & $0.591{\pm}0.004$ & $0.588{\pm}0.006$ & $0.586{\pm}0.007$ \\
 & best\_val & $0.999{\pm}0.001$ & $0.998{\pm}0.001$ & $0.999{\pm}0.001$ \\
\midrule
HOPE & Image-only & \multicolumn{3}{c}{$0.399$} \\
 & head & $0.459{\pm}0.014$ & $0.429{\pm}0.013$ & $0.431{\pm}0.016$ \\
 & fused & $0.421{\pm}0.003$ & $0.424{\pm}0.007$ & $0.423{\pm}0.006$ \\
 & best\_val & $1.000$ & $0.999{\pm}0.001$ & $0.999{\pm}0.002$ \\
\midrule
\fgip{} & Image-only & \multicolumn{3}{c}{$0.803$} \\
 & head & $0.904{\pm}0.026$ & $0.886{\pm}0.025$ & $0.884{\pm}0.023$ \\
 & fused & $0.860{\pm}0.007$ & $0.868{\pm}0.012$ & $0.873{\pm}0.010$ \\
 & best\_val & $1.000$ & $1.000$ & $1.000$ \\
\midrule
\multicolumn{5}{l}{Real-head / fused gains over image-only (percentage points)} \\
\tless{} & \multicolumn{4}{c}{$+0.5$ / $+3.1$} \\
HOPE & \multicolumn{4}{c}{$+6.0$ / $+2.2$} \\
\fgip{} & \multicolumn{4}{c}{$+10.1$ / $+5.6$} \\
\bottomrule
\end{tabular}
\end{table}

\begin{table}[htbp]
\centering
\caption{DINOv2-giant: real minus control, in percentage points with $95\%$ intervals. Permutations are paired by head seed; random-matrix intervals combine between-matrix and real-head seed variation ($n=5$, $4$ degrees of freedom).}
\label{tab:codebookdiff}
\setlength{\tabcolsep}{2pt}
\begin{tabular}{llcc}
\toprule
Dataset & Level & Real$-$random [95\%] & Real$-$permuted [95\%] \\
\midrule
\tless{} & head & $+0.6$ [$-1.3$, $+2.4$] & $+0.8$ [$-1.0$, $+2.7$] \\
 & fused & $+0.3$ [$-0.5$, $+1.1$] & $+0.5$ [$+0.1$, $+0.8$] \\
\midrule
HOPE & head & $+3.0$ [$+1.0$, $+5.0$] & $+2.8$ [$+0.8$, $+4.7$] \\
 & fused & $-0.3$ [$-1.1$, $+0.5$] & $-0.3$ [$-0.6$, $+0.1$] \\
\midrule
\fgip{} & head & $+1.8$ [$-2.2$, $+5.8$] & $+2.0$ [$-2.2$, $+6.1$] \\
 & fused & $-0.8$ [$-2.3$, $+0.7$] & $-1.3$ [$-2.3$, $-0.4$] \\
\bottomrule
\end{tabular}
\end{table}

On HOPE, the real head exceeds exact permutations by $2.8$ points
($t=3.9$); $23/25$ paired cells and all five permutation means favor real. Treating permutation
as the unit gives $[+1.9,+3.6]$ points, consistent with the seed-paired interval in
Table~\ref{tab:codebookdiff}. The permuted arm keeps every class-agnostic statistic of $G_s$, so
the HOPE head advantage comes from which class receives which row, consistent with the
class-semantic prototype arrangements of \citet{mettes2019hpn}. On \tless{} and \fgip{}, $17/25$
and $19/25$ cells favor the real head, the paired intervals include zero, and a head-level
difference is not detected. The HOPE residual does not reach fusion: real minus permuted
varies only from $-0.3$ to $+0.3$ points across the inspected weight grid.
Held-out onboarding accuracy is $0.999\pm0.001$ for real on \tless{} and $1.000$ on HOPE and
\fgip{}; control averages are $0.998$--$1.000$, with a minimum individual run of $0.993$
(HOPE, permuted). Every tested prototype matrix is therefore fitted; the differences on the queries concern transfer.

\paragraph{Prototype-matrix statistics and fusion weight}\label{sec:statistics}
Random prototype matrices are worse conditioned than the real one on \tless{} and better conditioned on
\fgip{}, and in neither case does head accuracy follow (\ref{app:statistics},
Fig.~\ref{fig:codebookstats}a).

On the grid of fusion weights scored on test labels, a diagnostic, mean accuracy is largest at
$\lambda=1$ for the real, random and permuted prototypes on all three datasets
(Fig.~\ref{fig:codebookstats}b). The best fused system on this grid leans hardest on the supervised
head whatever the prototype matrix. The weights calibrated on onboarding frames in
Section~\ref{sec:appearance} sit at the top of their grid for most HOPE rules, in its upper half
for most \tless{} rules, and are unstable on \fgip.

\subsection{Matched label supervision}\label{sec:supervision}
If the prototype matrix is a frozen output layer, what the scan contributes to the head is a labeled
support set on the same frozen features. The reference to beat is a classifier trained on that
support set with no geometry at all~\citep{tian2020rethinking}.
Table~\ref{tab:supervision} compares the audited system with geometry-free learners.
Matched logistic regression comes within $0.5$ points of fixed fusion on \tless{} and exceeds it
on HOPE and \fgip. The softmax reference trails logistic regression on \tless{} and HOPE but leads
on \fgip{}, showing that the choice of classifier matters. The \fgip{} softmax head also exceeds
the geometry head. Full-budget and
matched results differ by about two points or less. HOPE retains a head-level advantage of $2.6$ to $3.8$ points over the matched
references, comparable to the real-minus-permuted difference in Section~\ref{sec:codebook}.

\begin{table}[htbp]
\centering
\caption{Supervision-only references on DINOv2-giant (top-1, mean$\pm$SD over $5$ seeds where supplied). Matched uses the head optimization subset and, for softmax, its validation selection rule; full budget uses all onboarding frames and the final softmax epoch. W denotes white-cutout queries; other rows use real backgrounds.}
\label{tab:supervision}
\setlength{\tabcolsep}{3pt}
\begin{tabular}{llccc}
\toprule
Queries & Training / classifier & \tless{} & HOPE & \fgip{} \\
\midrule
Real & Matched LR & $0.586{\pm}0.002$ & $0.433{\pm}0.002$ & $0.916{\pm}0.002$ \\
 & Matched softmax & $0.529{\pm}0.006$ & $0.421{\pm}0.012$ & $0.947{\pm}0.004$ \\
 & Full LR & $0.585$ & $0.454$ & $0.921$ \\
 & Full softmax & $0.522{\pm}0.003$ & $0.409{\pm}0.007$ & $0.948{\pm}0.003$ \\
 & Geometry head & $0.566{\pm}0.015$ & $0.459{\pm}0.014$ & $0.904{\pm}0.026$ \\
 & Fused & $0.591{\pm}0.004$ & $0.421{\pm}0.003$ & $0.860{\pm}0.007$ \\
\midrule
W & Full LR & --- & $0.903$ & $0.901$ \\
 & Full softmax & --- & $0.900{\pm}0.007$ & $0.911{\pm}0.005$ \\
 & Geometry head & --- & $0.920{\pm}0.007$ & $0.912{\pm}0.009$ \\
 & Fused & --- & $0.872{\pm}0.002$ & $0.857{\pm}0.002$ \\
\bottomrule
\end{tabular}
\end{table}

\subsection{Stronger appearance matching}\label{sec:appearance}
% Figure 4 is declared ahead of the first paragraph so that it is set on the page after its first citation.
\begin{figure}[tbp]
\centering
\includegraphics[width=\linewidth]{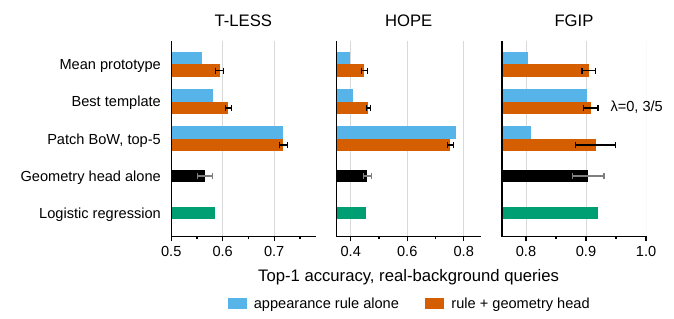}
\caption{Appearance rules with and without the geometry head on DINOv2-giant, real-background queries, for \tless{}, HOPE and \fgip{} (top-1 accuracy). Three rules are shown: mean prototype, best single template, and final-layer patch bag-of-words with best-five matching. Light blue bars give the rule alone. Orange bars give the rule fused with the real geometry head at a weight calibrated on held-out onboarding frames (mean and SD over $5$ head seeds). Black bars give the head alone and green bars full-budget logistic regression without geometry. The label ``$\lambda=0$, $3/5$'' marks a rule whose calibrated weight is zero in three of five seeds. Table~\ref{tab:appearancefull} lists all nine rules.}
\label{fig:appearance}
\end{figure}

Replacing the mean prototype by a stronger appearance rule changes both the reference level and
the gain the head can still add (Table~\ref{tab:appearancefull}; Fig.~\ref{fig:appearance}). On
\tless{}, global rules gain
$2.6$--$3.4$ points from the head, while final-layer patch best-five gains only $0.1$.
The final-layer patch matcher is a stronger appearance classifier than the global rules. Once it
is in place the supervised head has little left to supply, the same absorption that
Section~\ref{sec:supervision} shows with a linear probe.
On HOPE, global rules gain $4.4$--$5.5$ points with the real prototype matrix $G$ and $4.0$--$5.4$ with
random matrices (Table~\ref{tab:protocol}, last column), so this gain is the supervised head at work.
Patch best-five alone is $31.5$ points above the geometry head; adding that head loses
$2.1$ points. Even choosing its weight with test labels would gain only one of $915$ queries.
On HOPE the choice of appearance representation is worth far more than the geometry branch.
In patch fusion the real head sits $1.7$--$2.7$ points above the random-matrix heads, and both
lower the strongest patch baseline. The random heads have one seed per matrix and no interval, so
we report the direction and do not test it.

\begin{table}[htbp]
\centering
\caption{Appearance rules with real-background queries: alone and fused with the real geometry head (top-1; fused mean$\pm$SD, $5$ seeds). Each rule calibrates its fusion weight on held-out onboarding frames. HOPE $k$-means alone is $0.410$ at seed $0$ and ranges from $0.402$ to $0.423$. FGIP mean-prototype features were re-encoded, giving $0.804$ instead of $0.803$. Geometry-only and full-budget LR references are in Table~\ref{tab:supervision}.}
\label{tab:appearancefull}
\setlength{\tabcolsep}{3pt}
\begin{tabular}{>{\raggedright\arraybackslash}p{162pt}cc}
\toprule
\multicolumn{3}{l}{\tless{}} \\
Rule & Alone & $+$ head \\
\midrule
Mean prototype & $0.560$ & $0.594{\pm}0.008$ \\
Best template & $0.581$ & $0.611{\pm}0.006$ \\
Mean of best five & $0.584$ & $0.610{\pm}0.006$ \\
$k$-means ($8$ centers) & $0.578{\pm}0.001$ & $0.607{\pm}0.007$ \\
Soft vote, all templates & $0.575$ & $0.604{\pm}0.007$ \\
\addlinespace[2pt]
Patch, last, best & $0.702$ & $0.714{\pm}0.006$ \\
Patch, last, best five & $0.717$ & $0.718{\pm}0.008$ \\
Patch, L30, best & $0.628$ & $0.672{\pm}0.005$ \\
Patch, L30, best five & $0.642$ & $0.682{\pm}0.006$ \\
\midrule
\multicolumn{3}{l}{HOPE} \\
Rule & Alone & $+$ head \\
\midrule
Mean prototype & $0.399$ & $0.448{\pm}0.010$ \\
Best template & $0.408$ & $0.463{\pm}0.007$ \\
Mean of best five & $0.420$ & $0.468{\pm}0.005$ \\
$k$-means ($8$ centers) & $0.410$ & $0.457{\pm}0.007$ \\
Soft vote, all templates & $0.414$ & $0.458{\pm}0.009$ \\
\addlinespace[2pt]
Patch, last, best & $0.745$ & $0.732{\pm}0.010$ \\
Patch, last, best five & $0.774$ & $0.753{\pm}0.011$ \\
Patch, L30, best & $0.730$ & $0.730{\pm}0.007$ \\
Patch, L30, best five & $0.737$ & $0.735{\pm}0.007$ \\
\midrule
\multicolumn{3}{l}{\fgip{}} \\
Rule & Alone & $+$ head \\
\midrule
Mean prototype & $0.804$ & $0.905{\pm}0.011$ \\
Best template & $0.901$ & $0.908{\pm}0.012$ \\
Mean of best five & $0.893$ & $0.916{\pm}0.029$ \\
$k$-means ($8$ centers) & $0.841{\pm}0.004$ & $0.908{\pm}0.027$ \\
Soft vote, all templates & $0.866$ & $0.921{\pm}0.009$ \\
\addlinespace[2pt]
Patch, last, best & $0.790$ & $0.815{\pm}0.055$ \\
Patch, last, best five & $0.808$ & $0.916{\pm}0.033$ \\
Patch, L30, best & $0.774$ & $0.836{\pm}0.057$ \\
Patch, L30, best five & $0.820$ & $0.938{\pm}0.020$ \\
\bottomrule
\end{tabular}
\end{table}

On \fgip{}, best-template matching is within one point of the geometry head and gains
$0.7$ points from fusion; other rules gain $2.3$--$11.8$. Calibration is unstable: the selected
weight is zero in $3/5$ seeds for the global best-template rule and $4/5$ for the patch best-template
rule, with fused SDs reaching $0.06$. Full-budget logistic regression matches or exceeds all
fused values except $0.938$, and the softmax head exceeds all, so the \fgip{} fused values are
inconclusive. Patch best-five ($0.808$) also
trails the global best template ($0.901$): a stronger representation on one dataset need not be
stronger on another.

\subsection{Paired query backgrounds}\label{sec:protocol}
\begin{figure}[htbp]
\centering
\includegraphics[width=\linewidth]{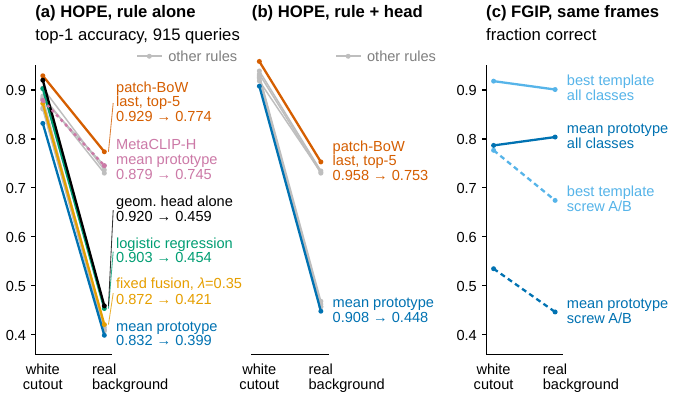}
\caption{Paired query protocols on DINOv2-giant features (top-1 accuracy). (a, b) The same $915$ HOPE query boxes scored as white cutouts and with the scene background kept. (a) Rules alone: the mean prototype (blue) and final-layer patch bag-of-words with best-five matching (vermilion) are labeled; gray lines are the other seven rules. References: full-budget logistic regression (green), head alone (black), fixed fusion at $\lambda=0.35$ (orange) and the MetaCLIP-H mean prototype (dotted purple). (b) The nine rules fused with the real geometry head at the onboarding-calibrated weight (mean of $5$ seeds). (c) \fgip{}, the same $1{,}927$ frames: mean-prototype and best-template rules on all classes (solid) and on the screw A/B pair (dashed).}
\label{fig:protocol}
\end{figure}

Table~\ref{tab:protocol} and Fig.~\ref{fig:protocol} score the same $915$ HOPE queries under the
white-cutout and real-background protocols of Section~\ref{sec:datasets}. Keeping the scene content
lowers DINOv2-giant
image-only accuracy by $43.3$ points and the other global rules by $45$--$47$ points. Patch
matching loses less ($13.6$--$16.7$ points). Fused with the geometry head, the strongest rule,
patch best-five, gains $2.9$ points on white cutouts and loses $2.1$ points with real backgrounds:
on HOPE the sign of the head's contribution to the strongest appearance rule depends on the
protocol. The contributions of background, neighboring objects and occluders are not separated.
The MetaCLIP-H mean-prototype drop is $13.3$ points, so the $43$-point effect belongs to the
DINOv2-giant global feature.

\begin{table}[htbp]
\centering
\caption{Paired query protocols. White pastes the masked foreground on a white canvas; Real keeps the scene content of the box (Section~\ref{sec:datasets}). HOPE uses the same $915$ boxes; calibrated fusion uses $5$ real-head seeds or $5$ random-matrix heads (one seed each). The HOPE white $k$-means result is $0.861\pm0.011$. FGIP parentheses count correct screw A/B queries. T-LESS uses real backgrounds throughout; its corresponding results are in Tables~\ref{tab:supervision} and~\ref{tab:appearancefull}.}
\label{tab:protocol}
\setlength{\tabcolsep}{2pt}
\begin{tabular}{>{\raggedright\arraybackslash}p{111pt}ccccc}
\toprule
& \multicolumn{2}{c}{Alone} & \multicolumn{2}{c}{$+$ real head} & $+$ random \\
\cmidrule(lr){2-3}\cmidrule(lr){4-5}
HOPE rule & White & Real & White & Real & Real \\
\midrule
Mean prototype & $0.832$ & $0.399$ & $0.908$ & $0.448$ & $0.440$ \\
MetaCLIP mean & $0.879$ & $0.745$ & --- & --- & --- \\
Best template & $0.863$ & $0.408$ & $0.925$ & $0.463$ & $0.461$ \\
Mean of best five & $0.882$ & $0.420$ & $0.930$ & $0.468$ & $0.468$ \\
$k$-means ($8$) & $0.861$ & $0.410$ & $0.924$ & $0.457$ & $0.458$ \\
Soft vote & $0.883$ & $0.414$ & $0.928$ & $0.458$ & $0.454$ \\
Patch, last, best & $0.881$ & $0.745$ & $0.918$ & $0.732$ & $0.715$ \\
Patch, last, five & $0.929$ & $0.774$ & $0.958$ & $0.753$ & $0.729$ \\
Patch, L30, best & $0.887$ & $0.730$ & $0.932$ & $0.730$ & $0.703$ \\
Patch, L30, five & $0.904$ & $0.737$ & $0.939$ & $0.735$ & $0.710$ \\
LR, full budget & $0.903$ & $0.454$ & --- & --- & --- \\
Head alone & $0.920$ & $0.459$ & --- & --- & --- \\
Fusion, $\lambda=.35$ & $0.872$ & $0.421$ & --- & --- & --- \\
\midrule
\multicolumn{6}{l}{\fgip{}: same $1{,}927$ queries; screws A/B subset $n=215$} \\
& \multicolumn{2}{c}{White} & \multicolumn{3}{c}{Real} \\
Mean, correct & \multicolumn{2}{c}{$1{,}516$ ($115$)} & \multicolumn{3}{c}{$1{,}549$ ($96$)} \\
Best, correct & \multicolumn{2}{c}{$1{,}769$ ($167$)} & \multicolumn{3}{c}{$1{,}736$ ($145$)} \\
Image-only & \multicolumn{2}{c}{$0.787$} & \multicolumn{3}{c}{$0.803$} \\
Head & \multicolumn{2}{c}{$0.912$} & \multicolumn{3}{c}{$0.904$} \\
Fused ($\lambda=.35$) & \multicolumn{2}{c}{$0.857$} & \multicolumn{3}{c}{$0.860$} \\
\bottomrule
\end{tabular}
\end{table}

On \fgip{}, the white cutout changes mean-prototype and best-template matching by $-33$ and $+33$
correct queries, respectively, and raises both on the $215$ screw A/B queries. The audited system
changes by less than two points in either
direction. The white cutout changes the task and can change method rankings: it simplifies the
evaluated HOPE scenes substantially, and its effect on \fgip{} is rule-dependent. Cropping at the
ground-truth box is one oracle step; replacing the background inside the box using the ground-truth
visibility mask is a second. Papers should state which one they use and report both when the second
is used.

\subsection{A second frozen backbone}\label{sec:backbone2}
\begin{figure}[htbp]
\centering
\includegraphics[width=\linewidth]{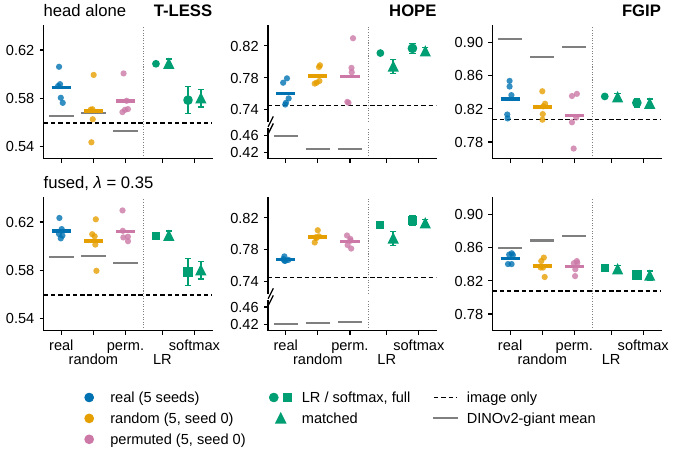}
\caption{Prototype controls on MetaCLIP-H with real-background queries, for \tless{}, HOPE and \fgip{} (columns; the HOPE axis is broken). Top row: top-1 accuracy of the head alone; bottom row: fused accuracy at $\lambda=0.35$. Each point is one training run and each bar the mean of its arm: the real prototype matrix (blue, $5$ head seeds), $5$ random prototype matrices (orange) and $5$ exact permutations $P_kG_0$ (purple), both at head seed $0$. Green markers right of the dotted line are geometry-free references on MetaCLIP-H features: logistic regression (LR) and softmax head trained on all onboarding frames (circle, square) and on the matched subset (triangles; mean and SD over $5$ seeds). The dashed line is MetaCLIP-H image-only accuracy. Short gray lines give the DINOv2-giant arm means under the same design.}
\label{fig:backbone2}
\end{figure}

MetaCLIP-H~\citep{xu2024metaclip} replaces CLS features with $1024$-D projected image features;
the saved real, random and permuted matrices are unchanged. Real uses five head seeds; random
and permuted arms use five random matrices and five permutations of $G_0$, each at head seed $0$.
The same-seed differences at seed $0$ are therefore the cleaner comparison.
Tables~\ref{tab:backbone2}--\ref{tab:backbone2refs} and Fig.~\ref{fig:backbone2} report this
design, its same-seed comparison, matched supervised references, and a DINOv2-giant repetition
under the same design.

\begin{table}[htbp]
\centering
\caption{MetaCLIP-H prototype audit (top-1, mean$\pm$SD, real backgrounds; $\lambda=0.35$). Real uses $5$ head seeds; random uses $5$ prototype matrices at head seed $0$; permuted uses $5$ exact permutations $P_kG_0$ at head seed $0$. Image-only gives MetaCLIP-H / DINOv2-giant; best\_val is held-out onboarding accuracy.}
\label{tab:backbone2}
\setlength{\tabcolsep}{2pt}
\begin{tabular}{llccc}
\toprule
Dataset & Level & Real & Random & Permuted \\
\midrule
\tless{} & image-only & \multicolumn{3}{c}{$0.559$ / $0.560$} \\
 & head & $0.589{\pm}0.012$ & $0.569{\pm}0.020$ & $0.578{\pm}0.013$ \\
 & fused & $0.613{\pm}0.007$ & $0.604{\pm}0.016$ & $0.612{\pm}0.010$ \\
 & best\_val & $0.995$ & $0.992$ & $0.992$ \\
\midrule
HOPE & image-only & \multicolumn{3}{c}{$0.745$ / $0.399$} \\
 & head & $0.760{\pm}0.015$ & $0.782{\pm}0.011$ & $0.781{\pm}0.034$ \\
 & fused & $0.768{\pm}0.002$ & $0.796{\pm}0.005$ & $0.790{\pm}0.006$ \\
 & best\_val & $1.000$ & $1.000$ & $0.999$ \\
\midrule
\fgip{} & image-only & \multicolumn{3}{c}{$0.807$ / $0.803$} \\
 & head & $0.832{\pm}0.020$ & $0.822{\pm}0.013$ & $0.812{\pm}0.027$ \\
 & fused & $0.847{\pm}0.006$ & $0.838{\pm}0.009$ & $0.837{\pm}0.007$ \\
 & best\_val & $0.999$ & $1.000$ & $1.000$ \\
\bottomrule
\end{tabular}
\end{table}

\begin{table}[htbp]
\centering
\caption{MetaCLIP-H: real minus control in percentage points. Run comparisons give Welch $t$ in parentheses; the last two columns match head seed $0$.}
\label{tab:backbone2diff}
\setlength{\tabcolsep}{2pt}
\begin{tabular}{llcccc}
\toprule
& & \multicolumn{2}{c}{Real minus, over runs ($t$)} & \multicolumn{2}{c}{Real minus, seed $0$} \\
\cmidrule(lr){3-4}\cmidrule(lr){5-6}
Dataset & Level & random & permuted & random & permuted \\
\midrule
\tless{} & head & $+2.0$ ($1.9$) & $+1.1$ ($1.4$) & $+1.1$ & $+0.3$ \\
 & fused & $+0.8$ ($1.1$) & $+0.0$ ($0.0$) & $+0.5$ & $-0.4$ \\
\midrule
HOPE & head & $-2.2$ ($-2.6$) & $-2.1$ ($-1.3$) & $-0.9$ & $-0.8$ \\
 & fused & $-2.8$ ($-10.6$) & $-2.2$ ($-7.2$) & $-3.0$ & $-2.4$ \\
\midrule
\fgip{} & head & $+1.0$ ($0.9$) & $+2.0$ ($1.3$) & $-1.4$ & $-0.3$ \\
 & fused & $+1.0$ ($1.9$) & $+1.0$ ($2.2$) & $+0.2$ & $+0.3$ \\
\bottomrule
\end{tabular}
\end{table}

\begin{table}[htbp]
\centering
\caption{Matched references on MetaCLIP-H (mean$\pm$SD, $5$ seeds) and the second-backbone control design applied to DINOv2-giant. DINO real uses $5$ head seeds; random and exact permutations use head seed $0$. All queries retain their backgrounds; fusion uses $\lambda=0.35$.}
\label{tab:backbone2refs}
\setlength{\tabcolsep}{3pt}
\begin{tabular}{>{\raggedright\arraybackslash}p{132pt}ccc}
\toprule
Reference / arm & \tless{} & HOPE & \fgip{} \\
\midrule
\multicolumn{4}{l}{MetaCLIP-H, matched optimization subset} \\
Logistic regression & $0.609{\pm}0.004$ & $0.794{\pm}0.009$ & $0.834{\pm}0.005$ \\
Softmax head & $0.580{\pm}0.007$ & $0.813{\pm}0.005$ & $0.826{\pm}0.006$ \\
Geometry head & $0.589{\pm}0.012$ & $0.760{\pm}0.015$ & $0.832{\pm}0.020$ \\
\midrule
\multicolumn{4}{l}{DINOv2-giant: head / fused} \\
Real ($5$ seeds) & $0.566$ / $0.591$ & $0.459$ / $0.421$ & $0.904$ / $0.860$ \\
Random (seed $0$) & $0.568$ / $0.591$ & $0.428$ / $0.423$ & $0.883$ / $0.869$ \\
Permuted (seed $0$) & $0.553$ / $0.586$ & $0.427$ / $0.426$ & $0.894$ / $0.874$ \\
\bottomrule
\end{tabular}
\end{table}

The real prototype matrix improves fusion over either control by at most one point. On HOPE, every random and permuted fused run
exceeds every real fused run; the fused deficit remains when matching head seed $0$.
The HOPE head-level advantage observed with DINOv2-giant was not reproduced in the tested
MetaCLIP-H configuration. Across datasets, same-seed head differences remain within $1.4$ points
(Table~\ref{tab:backbone2diff}). Held-out arm means range from $0.992$ to $1.000$
(individual runs: $0.989$--$1.000$). Matched logistic regression is at or above the geometry head
on all three datasets, and the HOPE softmax head exceeds both, the geometry head by $5.2$ points.
MetaCLIP-H image-only accuracy on real-background HOPE is $0.745$, against $0.774$ for the DINOv2-giant patch
matcher.

\subsection{Reconstruction source, descriptor, image encoder and rendered pixels}\label{sec:extensions}
Four measurements on \tless{} each vary one input of the audited system
(\ref{app:extensions}). Prototype matrices built from onboarding depth, from the \dgs{}
reconstruction and from surface-sampled CAD agree within $1.6$ points at the head and $0.5$ points
fused. An in-domain PointNet++ and a frozen OpenShape encoder~\citep{liu2023openshape}, each in
place of the handbuilt descriptor, stay within its seed variability. Random and permuted prototype
matrices also stay within $0.5$ points of the real fused accuracy on \tless{}
(Section~\ref{sec:codebook}), the same range as the source and descriptor changes. Among $14$
frozen image encoders, fusion at
$\lambda=0.35$ improves on the image prototype for $13$ (Table~\ref{tab:backbone}), and the gain
does not track image-only strength. Replacing the real onboarding images by \dgs{} renders lowers
image-only accuracy from $0.560$ to $0.392$; adding renders to them yields $0.554$. A CAD-lighting
probe changes accuracy by at most $2.5$ points. Rendered pixels do not help this mean-prototype,
frozen-DINOv2 pipeline.

\subsection{Direct shape matching}\label{sec:conditions}
To distinguish the result for the fixed prototype matrix $G$ from direct geometric matching, a
separate pilot gives the query a 3D measurement. On six
confusable \tless{} pairs ($273$ queries), two candidate references (CAD, \dgs{} scan or
onboarding depth) are registered to the query depth inside the visibility mask, and a
depth and free-space cost decides between the two objects of the pair. The pilot is post hoc and
its inputs are privileged: the ground-truth mask in every geometry arm and the ground-truth pose
in the pose-assisted arms. With $24$ PCA initializations and ICP,
object-balanced two-way accuracy is $0.486$--$0.503$, against $0.671$ for the raw-crop appearance
matcher. The pose-assisted arms, which initialize both candidates from the true identity's pose,
give $0.774$--$0.833$ (Table~\ref{tab:pilot}). On two pairs with $87$ usable queries, the true candidate
was displaced from its ground-truth pose until the decision flipped. The median first flip is at
$1.0$\,mm in translation and $6^{\circ}$ in rotation under shared-pose placement
(Table~\ref{tab:pilot}); $1.0$\,mm is the first grid point, and queries already wrong at zero
displacement are included. \ref{app:conditions} gives the
placement definitions, the perturbation grid, and the visibility and silhouette diagnostics.

\subsection{External semantic-prototype experiment}
\label{sec:hpn}

We apply the assignment intervention to Hyperspherical Prototype Networks (HPN)~\citep{mettes2019hpn},
a published recognizer with fixed semantic prototypes. The experiment uses the HPN authors' ResNet-32
and their $100$ ten-dimensional prototypes on the official CIFAR-100 split ($50{,}000$ training and
$10{,}000$ test images). Each control reassigns the same prototype rows by a derangement, which keeps the row set
and the Gram matrix up to permutation. The whole network is trained from scratch for $250$ epochs with
the HPN loss $\sum_i[1-\cos(z_i,g_{y_i})]^2$, where $z_i$ is the network output for image $i$, under
the native random crops and horizontal flips. Optimization uses SGD (learning rate $0.01$, momentum
$0.9$, weight decay $10^{-4}$), batches of $128$, and a tenfold learning-rate drop after epochs $100$
and $200$. Both arms
use PyTorch 2.8 with TF32 convolutions, and the final epoch is evaluated without test-based selection.
The semantic arm reaches $0.571{\pm}0.003$ test accuracy, in line with the $0.570{\pm}0.006$ that
\citet{mettes2019hpn} report for this setting.

The design crosses five training seeds with five fixed permutations: five semantic runs and $25$
permuted runs. Within each training seed the six runs share the initialization, the sample order and
every augmentation draw, which the stored run records verify through the initial-state hash and a
per-epoch hash of the data stream. The design follows one paired pilot, whose two runs are included
once; a sensitivity analysis drops the pilot seed and the pilot permutation and keeps four
seeds and four permutations.

\begin{table}[htbp]
\centering
\caption{HPN on CIFAR-100: top-1 accuracy of the semantic class assignment and of its derangements.
Test entries are mean$\pm$SD over five training seeds; permuted runs are first averaged over the five
permutations within a seed. Train: unaugmented training-set accuracy at the final epoch.}
\label{tab:hpn}
\begin{tabular}{lcc}
\toprule
Assignment (runs) & Test & Train \\
\midrule
Semantic ($5$) & $0.571{\pm}0.003$ & $0.822$ \\
Permuted ($5\times5$) & $0.500{\pm}0.003$ & $0.789$ \\
\bottomrule
\end{tabular}
\end{table}

Table~\ref{tab:hpn} summarizes the results. Semantic minus permuted is $+7.1$ points on the test set,
with a $95\%$ $t$-interval of $[+6.4,+7.8]$ over the five seed-level paired differences (four degrees
of freedom), conditional on the five permutations. The per-permutation means range from $+6.4$ to $+8.0$
points, and all $25$ cells are positive. Without the pilot seed and permutation the difference
is $+7.3$ points, $[+6.4,+8.2]$ (three degrees of freedom). The training-set difference is $+3.3$
points, so the two arms differ in fit as well as in generalization. Under this training budget the
published class assignment is worth $7$ points of test accuracy over derangements of the same
prototypes in HPN; the experiment compares assignments only and includes no supervised classifier
without prototypes.

\section{Discussion}\label{sec:discuss}

\paragraph{Attribution}
The controls show why improved accuracy over image-prototype matching alone cannot identify a
contribution from scanned shape. The onboarding labels support competitive geometry-free
classifiers, while arbitrary prototype matrices retain most of the fused benefit. The HOPE head-level
residual shows that class assignment can nevertheless affect learning; its value to the audited
recognizer must be assessed at the fused output. The reversal of the head's contribution under paired HOPE
query protocols further ties that value to the appearance representation and query processing.
For direct geometric matching, the depth pilot instead points to registration and visibility as
conditions requiring separate evaluation (Section~\ref{sec:conditions}).

\paragraph{A reproducible audit}
The interventions of Section~\ref{sec:exp} form an audit for fixed reference-derived
class-prototype matrices read by a class-supervised head. First identify where the reference enters the
computation. For a fixed class-prototype matrix,
freeze the queries, features, split and fusion rule. Compare the real matrix with independently
generated matrices passed through the same preprocessing and with exact fixed-point-free row
permutations of each saved real matrix. Pair the latter with the real arm's training seed and
split. Fit geometry-free linear and softmax classifiers on the same optimization subset; use the
same validation checkpoint rule for trainable heads. Apply the permutations also to directly read
image prototypes as a positive control. Repeat the audit on a second frozen backbone and compare
against a stronger appearance matcher, calibrating any new fusion weight only on held-out
onboarding frames. Report head, fused and onboarding accuracies; separate training seeds from
draws of the random prototype matrix; give paired or matrix-level differences with intervals. Pair alternative query
protocols image by image. Release prototype matrices, permutations, seeds, split indices and predictions.

\paragraph{What remains open}
Three questions remain open. The first is whether scanned geometry adds recognition evidence
when the query carries a 3D measurement and its pose is estimated (Section~\ref{sec:conditions}).
The second is which property of the HOPE classes gives the real class assignment its head-level
advantage on DINOv2-giant, and whether the advantage appears on MetaCLIP-H with more training seeds
(Sections~\ref{sec:codebook} and~\ref{sec:backbone2}). The third is whether the $1.7$--$2.7$ point
difference between real and random heads in patch fusion holds with several seeds per random
prototype matrix (Section~\ref{sec:appearance}).

\paragraph{Scope}
Segmented onboarding images and real-background queries create a domain gap whose contribution
was not separately measured. The scan-based evidence spans $74$ objects in three domains, two image backbones
for the controls and a closed-set few-shot setting. HPN separately tests class assignment with
semantic prototypes and full-network training; differences between the two systems do not isolate
which property makes an assignment useful.

\section{Conclusion}\label{sec:conclusion}
This study separates the contribution of scan-derived class prototypes from label supervision,
appearance strength and query protocol in a recognizer over frozen features. Across three
datasets and two backbones, the largest fused-accuracy advantage of the real prototypes over either
control is one percentage point in the observed arm means. Their conditional head-level advantage
on HOPE with DINOv2-giant does not improve the audited fusion. In HPN on CIFAR-100, the published class assignment is $7$
points above derangements of the same prototypes under one training budget. Matched supervision and paired
query protocols provide the comparisons needed to interpret these outcomes. For systems using
fixed class-prototype matrices and class-supervised heads, the audit tests whether gains depend on the
original shape-to-class assignment and improve the final prediction.

\section*{CRediT authorship contribution statement}
\textbf{Chenxi Tao:} Conceptualization, Methodology, Software, Formal analysis, Investigation, Data curation, Writing -- original draft, Writing -- review \& editing, Visualization. \textbf{Hong-In Won:} Funding acquisition, Writing -- review \& editing. \textbf{Seung-Kyum Choi:} Supervision, Funding acquisition, Conceptualization, Writing -- review \& editing.

\section*{Declaration of competing interest}
The authors declare that they have no known competing financial interests or personal relationships that could have appeared to influence the work reported in this paper.

\section*{Data availability}
T-LESS and HOPE are available through the BOP benchmark (\url{https://bop.felk.cvut.cz}), and
CIFAR-100 at \url{https://www.cs.toronto.edu/~kriz/cifar.html}. The reproducibility package,
which contains the source-linked numerical ledger, the control-generation and run scripts, the $75$
exact-permutation matrices and indices, the matched split indices, the per-query predictions for the
permutation, MetaCLIP and reference runs, the $915$-query real-background HOPE index, and the HPN
prototype provenance, scripts, permutation indices, run records and test predictions, is available
from the corresponding author on request, as are the FGIP crops, reconstructions, prototype matrices
and analysis code.

\section*{Funding}
This research was funded by the Ministry of Trade, Industry and Energy (MOTIE), Korea, through the Global Industrial Technology Cooperation Center (GITCC) Program, supervised by the Korea Institute for Advancement of Technology (KIAT) (Task No. P24680172) and by the Manufacturing AI Key-Tech Program, supervised by the Korea Institute of Industrial Technology (P28726). The funding sources had no involvement in study design; in the collection, analysis and interpretation of data; in the writing of the report; or in the decision to submit the article for publication.

\section*{Acknowledgements}
The authors thank the GITCC Program (MOTIE/KIAT) for its support, and gratefully acknowledge the maintainers of the \tless{} and HOPE datasets and the BOP benchmark, and the open-source Nerfstudio and DINOv2 projects, whose resources made this study possible.

\appendix
% elsarticle numbers appendix floats as A.1, B.1, ... but does not reset the counters itself
\makeatletter
\@addtoreset{figure}{section}
\@addtoreset{table}{section}
\makeatother
\setcounter{figure}{0}
\setcounter{table}{0}
\renewcommand{\theHfigure}{\Alph{section}.\arabic{figure}}
\renewcommand{\theHtable}{\Alph{section}.\arabic{table}}

\section{Prototype-matrix statistics and fusion weight}\label{app:statistics}
\begin{figure}[htbp]
\centering
\includegraphics[width=\linewidth]{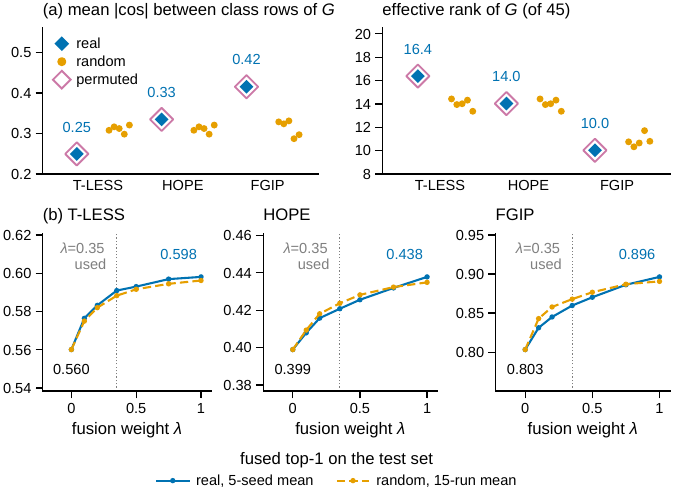}
\caption{Prototype-matrix diagnostics on \tless{}, HOPE and \fgip{}. (a) Mean absolute cosine between class rows of $G$ (left) and effective rank of $G$ out of $45$ (right): real prototype matrix (blue diamond, value labeled), five random prototype matrices (orange points) and exact permutations (open purple diamond, identical to real by construction). \tless{} and HOPE share the random matrices. (b) Fused top-1 accuracy on the test queries (DINOv2-giant, real backgrounds) against the fusion weight $\lambda$: mean over $5$ real-matrix runs (blue, solid) and $15$ random-matrix runs (orange, dashed). Labels give image-only accuracy at $\lambda=0$ and the real-matrix value at $\lambda=1$. The dotted line marks $\lambda=0.35$; the grid is scored on test labels as a diagnostic.}
\label{fig:codebookstats}
\end{figure}

% Inside the appendices a pending float does not force a page break at the next appendix
% heading (placeins [section] would leave most of a page empty); the barrier returns before the references.
\let\mainFloatBarrier\FloatBarrier
\renewcommand{\FloatBarrier}{\par}
Figure~\ref{fig:codebookstats}a compares mean absolute row cosine and effective rank.
For singular values $\sigma_j$, the latter is $(\sum_j\sigma_j)^2/\sum_j\sigma_j^2$.
Real prototype matrices give $(0.249,16.4)$ on \tless{},
$(0.335,14.0)$ on HOPE and $(0.415,10.0)$ on \fgip{}. Random-matrix averages are
$(0.311,14.0)$ for the $28$-class sets and $(0.313,10.8)$ for \fgip{}.
Permutations preserve these statistics; the row-Gram identity has a maximum floating-point
discrepancy of $3.6\times10^{-7}$ over the $75$ saved matrices.

Across the $15$ random runs, the SD of the five matrix-level means is $0.003$, $0.009$, $0.019$
on \tless{}, HOPE, \fgip{}; the SD of the three seed means is $0.007$, $0.002$, $0.004$.
A descriptive two-way decomposition assigns $12\%/38\%$, $42\%/2\%$, $48\%/1\%$ of the
sum of squares to matrix/seed, respectively; residuals account for the rest ($F\le2.0$ for
the matrix factor). The random arm is therefore reported with the prototype matrix as the unit
(Section~\ref{sec:setup}).

At $\lambda=1$ on the test-label grid of Fig.~\ref{fig:codebookstats}b, real fusion reaches
$0.598$, $0.438$, $0.896$ ($+0.7$, $+1.7$, $+3.7$ points over fixed fusion); random reaches
$0.596$, $0.435$, $0.891$; permuted reaches $0.594$, $0.435$, $0.889$. The score at $\lambda=1$
still includes the image branch. The HOPE head classifies all $560$ held-out frames correctly,
which drives the onboarding calibration to its upper grid bound.

\section{Reconstruction source, descriptor, image encoder and rendered pixels}\label{app:extensions}
\begin{table}[htbp]
\centering
\caption{T-LESS frozen-backbone comparison, sorted by fused accuracy ($\lambda=0.35$; mean$\pm$SD over $5$ seeds; $^{\dagger}n=4$). Image-only is deterministic. SDs in this table use the divisor $n$; Tables~\ref{tab:codebook} and~\ref{tab:backbone2} use $n-1$, which changes the last digit. Feature families and checkpoints are listed below the table.}
\label{tab:backbone}
\setlength{\tabcolsep}{3pt}
\begin{tabular}{lccc}
\toprule
Encoder & Image-only & Geometry head & Fused \\
\midrule
SigLIP2-g$^{\dagger}$ & $0.561$ & $0.611{\pm}0.010$ & $0.629{\pm}0.004$ \\
MetaCLIP-H & $0.559$ & $0.589{\pm}0.010$ & $0.613{\pm}0.006$ \\
SigLIP-L & $0.545$ & $0.555{\pm}0.002$ & $0.592{\pm}0.005$ \\
DINOv2-giant & $0.560$ & $0.566{\pm}0.013$ & $0.591{\pm}0.004$ \\
SigLIP2-so400m & $0.531$ & $0.553{\pm}0.013$ & $0.586{\pm}0.007$ \\
OpenCLIP-H & $0.546$ & $0.516{\pm}0.016$ & $0.578{\pm}0.011$ \\
DINOv3-L & $0.529$ & $0.557{\pm}0.010$ & $0.569{\pm}0.005$ \\
DINOv2-reg-g & $0.522$ & $0.546{\pm}0.011$ & $0.559{\pm}0.006$ \\
OpenCLIP-bigG & $0.548$ & $0.477{\pm}0.027$ & $0.556{\pm}0.012$ \\
SigLIP-so400m & $0.556$ & $0.490{\pm}0.019$ & $0.553{\pm}0.010$ \\
DINOv3-H+ & $0.493$ & $0.528{\pm}0.025$ & $0.550{\pm}0.006$ \\
DINOv2-reg-L & $0.478$ & $0.530{\pm}0.016$ & $0.528{\pm}0.005$ \\
CLIP-L & $0.316$ & $0.336{\pm}0.034$ & $0.361{\pm}0.019$ \\
I-JEPA-H & $0.118$ & $0.134{\pm}0.021$ & $0.152{\pm}0.020$ \\
\bottomrule
\end{tabular}
\par\vspace{4pt}
\begin{minipage}{\linewidth}\footnotesize\raggedright
Families: CLIP, OpenCLIP, MetaCLIP and SigLIP variants are image--text encoders; DINO variants are self-supervised transformers; I-JEPA is a joint-embedding predictive architecture.
Checkpoints, each name appended to its repository prefix. Under \texttt{facebook/\allowbreak }:
\texttt{dinov2-\allowbreak giant},
\texttt{metaclip-\allowbreak h14-\allowbreak fullcc2.5b},
\texttt{dinov3-\allowbreak vith16plus-\allowbreak pretrain-\allowbreak lvd1689m},
\texttt{dinov3-\allowbreak vitl16-\allowbreak pretrain-\allowbreak lvd1689m},
\texttt{dinov2-\allowbreak with-\allowbreak registers-\allowbreak giant},
\texttt{dinov2-\allowbreak with-\allowbreak registers-\allowbreak large}, and
\texttt{ijepa\_vith14\_1k}.
Under \texttt{google/\allowbreak }:
\texttt{siglip2-\allowbreak giant-\allowbreak opt-\allowbreak patch16-\allowbreak 384},
\texttt{siglip2-\allowbreak so400m-\allowbreak patch16-\allowbreak 384},
\texttt{siglip-\allowbreak so400m-\allowbreak patch14-\allowbreak 384}, and
\texttt{siglip-\allowbreak large-\allowbreak patch16-\allowbreak 256}.
Under \texttt{laion/\allowbreak }:
\texttt{CLIP-\allowbreak ViT-\allowbreak H-\allowbreak 14-\allowbreak laion2B-\allowbreak s32B-\allowbreak b79K} and
\texttt{CLIP-\allowbreak ViT-\allowbreak bigG-\allowbreak 14-\allowbreak laion2B-\allowbreak 39B-\allowbreak b160k}.
The remaining checkpoint is \texttt{openai/\allowbreak clip-\allowbreak vit-\allowbreak large-\allowbreak patch14}.
\end{minipage}
\end{table}

\paragraph{Reconstruction source}
On \tless{}, back-projected onboarding depth yields head/fused accuracies of $0.575{\pm}0.006$ and
$0.593{\pm}0.002$ (three seeds); surface-sampled CAD gives $0.582{\pm}0.020$ and $0.588{\pm}0.005$.
With the \dgs{} result in Table~\ref{tab:codebook}, the spread is $1.6$ points at the head and
$0.5$ fused.

\paragraph{Descriptor}
An in-domain PointNet++ trained on the reconstructed clouds with rotation, jitter, scale and
partial-view augmentation gives head/fused accuracies of $0.571{\pm}0.009$ and
$0.588{\pm}0.002$. A frozen OpenShape encoder~\citep{liu2023openshape}, using Point-BERT aligned
to OpenCLIP~\citep{yu2022pointbert}, gives $0.579{\pm}0.017$ and $0.592{\pm}0.007$ zero-shot.
Both use three seeds and lie within the variability of the handbuilt descriptor.

\paragraph{Image encoder}
Table~\ref{tab:backbone} compares $14$ image encoders on the same $4{,}480$ labeled onboarding
crops, real-background queries and fusion weight $\lambda=0.35$: five seeds, except SigLIP2-g (four, one
failure). The one encoder without a fusion gain is SigLIP-so400m, within seed variability. Gains of $6.8$, $3.1$ and $0.8$ points for SigLIP2-g, DINOv2-giant and
OpenCLIP-bigG do not track image-only strength. The head alone trails the image prototype for
OpenCLIP-H, OpenCLIP-bigG and SigLIP-so400m. AIMv2~\citep{fini2024aimv2} is omitted
because its features gave chance-level accuracy in this pipeline. Register variants and I-JEPA
follow \citet{darcet2024registers} and \citet{assran2023ijepa}.

\paragraph{Rendered pixels}
The replacement run computes the mean image prototypes from \dgs{} renders of each reconstruction
instead of the onboarding crops, and the addition run computes them from the crops and the renders
together. Both score the same $6{,}395$ queries with frozen DINOv2-giant and give $0.392$ and
$0.554$, against $0.560$ for the crops alone. The CAD-lighting probe uses $28$ objects, $24$ rendered views per object, matched/shifted/extreme
illumination and $1$, $5$ or $10$ support views. Its largest change is $2.5$ points ($0.947$ to
$0.922$, ten views, extreme illumination).

\section{Direct shape matching: depth, pose and silhouette diagnostics}\label{app:conditions}

\paragraph{Measured depth}
Six non-overlapping \tless{} pairs ($25/26$, $7/8$, $1/2$, $21/22$, $11/12$, $27/29$) were chosen
from onboarding patch-BoW center similarity before query scoring. We froze $273$ queries from
$19$ scenes, counting $15$ with unusable depth as wrong. Depth was back-projected with the camera
intrinsics and ground-truth visibility mask. Each candidate (CAD surface, cleaned \dgs{} scan
or same-source onboarding depth fusion) was registered and scored with depth and free-space
costs. Table~\ref{tab:pilot} lists every arm; the CAD-box control gives $0.644$ with the
ground-truth pose. The scan and CAD have $225$ and $224$ correct queries, respectively; their
ordering reverses between pairs $25/26$ and $21/22$.

\begin{table}[htbp]
\centering
\caption{Measured-depth pilot on T-LESS: object-balanced two-way accuracy, $273$ development queries across six pairs; $15$ unusable-depth queries count as wrong. Geometry receives the true visibility mask; pose-assisted arms also receive the true identity pose. RGB receives only the raw crop. Perturbation summaries use $87$ usable-depth queries from pairs $25/26$ and $21/22$.}
\label{tab:pilot}
\setlength{\tabcolsep}{2pt}
\begin{tabular}{>{\raggedright\arraybackslash}p{182pt}cc}
\toprule
Arm & Pose-free & Pose-assisted \\
\midrule
RGB patch BoW & $0.671$ & --- \\
CAD surface + query depth & $0.486$ & $0.824$ \\
\dgs{} scan + query depth & $0.492$ & $0.833$ \\
Onboarding depth + query depth & $0.503$ & $0.774$ \\
CAD box + query depth & $0.486$ & $0.644$ \\
\midrule
\multicolumn{3}{l}{Pair details: raw RGB (left); assisted CAD / \dgs{} (right)} \\
Pair $25/26$ ($44$ queries) & $0.650$ & $0.721$ / $0.938$ \\
Pair $21/22$ ($48$ queries) & $0.562$ & $0.625$ / $0.479$ \\
\midrule
\multicolumn{3}{l}{First flipping perturbation: median [Q1, Q3], $n=87$} \\
Translation, shared-pose placement & \multicolumn{2}{c}{$1.0$ mm [$1.0$, $2.0$]} \\
Translation, identity placement & \multicolumn{2}{c}{$1.0$ mm [$1.0$, $1.0$]} \\
Rotation, shared-pose placement & \multicolumn{2}{c}{$6^\circ$ [$3$, $6$]} \\
\midrule
\multicolumn{3}{l}{Pose-free placement error of the true candidate} \\
Mean-NN surface distance, median & \multicolumn{2}{c}{$4.1$--$4.8$ mm} \\
\bottomrule
\end{tabular}
\end{table}

\paragraph{Pose and visibility}
Pairs $25/26$ and $21/22$ share outer shells with local differences ($92$ queries; $87$ usable).
In shared-pose placement the true candidate sits at its ground-truth pose and the alternative
candidate at that pose composed with a fixed rigid alignment between the two objects'
onboarding-depth references; identity placement puts both candidates at the ground-truth pose.
Only the true candidate is perturbed.
Using CAD surfaces and an L1 scorer, we displaced the true candidate from its ground-truth pose
and recorded the first error along the most fragile tested direction. The translation grid is
$1/2/4/7$\,mm and the rotation grid $3/6/10^{\circ}$. Table~\ref{tab:pilot} gives the first-flip
medians and quartiles; the median rotation under identity placement is $3^{\circ}$. Pair $21/22$
alone has $45$ usable queries,
an identity-placement translation IQR of $0.0$--$1.0$\,mm and median rotation $3^{\circ}$.
Errors at zero displacement count as zero: $16/87$ shared-pose and $20/87$ identity-placement
queries, including $11/45$ and $12/45$ on pair $21/22$.

Pose-free median mean-nearest-neighbor surface distances for the true candidate are $4.1$--$4.8$\,mm; the coarse
point-pair-feature stage covers $9$--$11$\,mm. These use a different error definition from the
first-flip perturbation. Pose sensitivity, visibility, initial errors and scorer behavior are not
separated. Under identity placement, the $20$ incorrect CAD cases among $87$ queries expose
little of the distinguishing region (defined at $2$\,mm): the median valid-depth-pixel count is
$5.5$, versus a first quartile of $18$ among correct cases (one-sided Mann--Whitney
$p=2.5\times10^{-5}$; queries within a scene are dependent).

\paragraph{Silhouette overlap}
On the confusable \fgip{} screw pair under white cutouts, 2D silhouette overlap gives $176$
correct queries out of $215$, against $167$ for appearance matching. Over all objects it gives
$1{,}294$ of $1{,}927$, against $1{,}769$.

\let\FloatBarrier\mainFloatBarrier
\section*{Declaration of generative AI and AI-assisted technologies in the writing process}
During the preparation of this work the authors used OpenAI Codex and Anthropic Claude in order to
organize, draft and edit the manuscript text. After using these tools, the authors reviewed and edited
the content as needed and take full responsibility for the content of the publication.

\bibliographystyle{elsarticle-harv}
\bibliography{references}

\end{document}